\documentclass{article}

\PassOptionsToPackage{numbers}{natbib}

     \usepackage[preprint]{neurips_2019}

\usepackage[utf8]{inputenc} 
\usepackage[T1]{fontenc}    
\usepackage{hyperref}       
\usepackage{url}            
\usepackage{booktabs}       
\usepackage{amsfonts}       
\usepackage{nicefrac}       
\usepackage{microtype}      
\usepackage{graphicx}
\usepackage{multirow}
\usepackage{subfigure}
\usepackage{caption}
\usepackage{amsmath}

\newcommand{\be}{\begin{equation}}
\newcommand{\ee}{\end{equation}}
\newcommand{\mb}{\mathbf}

\title{Out-Of-Distribution Detection With Subspace Techniques And Probabilistic Modeling Of Features}

\author{
  Ibrahima Ndiour \\
  Intel Labs \\
  \And
  Nilesh. A. Ahuja \\ 
  Intel Labs\\
  \And
  Omesh Tickoo \\
  Intel Labs \\
}

\begin{document}
\maketitle

\begin{abstract}
	This paper presents a principled approach for detecting out-of-distribution (OOD) samples in deep neural networks (DNN). Modeling probability distributions on deep features has recently emerged as an effective, yet computationally cheap method to detect OOD samples in DNN. However, the features produced by a DNN at any given layer do not fully occupy the corresponding high-dimensional feature space. We apply linear statistical dimensionality reduction techniques and nonlinear manifold-learning techniques on the high-dimensional features in order to capture the true subspace spanned by the features. We hypothesize that such  lower-dimensional feature embeddings can mitigate the \emph{curse of dimensionality}, and enhance any  feature-based method for more efficient and effective performance. 
	In the context of uncertainty estimation and OOD, we show that the log-likelihood score obtained from the distributions learnt on this lower-dimensional subspace is more discriminative for OOD detection. We also show that the feature reconstruction error, which is the $L_2$-norm of the difference between the original feature and the pre-image of its embedding, is highly effective for OOD detection and in some cases superior to the log-likelihood scores. The benefits of our approach are demonstrated on image features by detecting OOD images, using popular DNN architectures on commonly used image datasets such as CIFAR10, CIFAR100, and SVHN.
\end{abstract}

\section{Introduction}
\label{sec:intro}
Deep networks deployed in real-world conditions will inevitably encounter out-of-distribution (OOD) data: data that is beyond that with which the network was trained. Networks should be able to indicate if the data that they are processing is OOD. This is a critical requirement for perceptual sub-systems based on deep learning if we are to build safe and transparent systems that do not adversely impact humans (e.g. in fields such as autonomous driving, robotics, or healthcare). Additional imperatives for estimating predictive uncertainty measures relate to results interpretability, dataset bias, AI safety, and active learning.

OOD detection is typically performed by making the network provide a confidence score for each input, with the intuition that in-distribution samples should be identified with high confidence while OOD samples should have low confidence. A variety of confidence measures have been studied in the literature. The softmax score \cite{hendrycks2016baseline} and its temperature-scaled variants \cite{liang2018enhancing} have been applied for OOD detection. This problem has also been studied in the context of Bayesian neural networks \cite{gal2016dropout, kendall2017uncertainties}, in which the network's parameters are represented by probability distributions rather than single point values. At inference, a predictive distribution over the outputs is generated by performing multiple stochastic forward passes. Uncertainty measures such as predictive mean and predictive entropy can be calculated, but this comes at the cost of significantly higher compute and memory cost. Similar statistics can be calculated from ensembles of discriminative classifiers \cite{lakshminarayanan2017simple}, but these also entail a high cost in terms of maintaining multiple networks for the ensemble. 

Another class of methods use generative modeling to learn distributions over the input data, and then evaluate the likelihood of new inputs with respect to the learnt distributions. Deep generative models such as PixelCNN \cite{van2016conditional} have been used for this purpose. \cite{hendrycks2019oe} investigated failure modes of this approach and proposed the method of Outlier Exposure to fix those problems, which involves training the model with an auxiliary out-of-distribution dataset which is different from the test data. \cite{ren2019likelihood} proposed a likelihood-ratio method for deep-generative models that outperformed the likelihood from such models. 

An alternative to learning a deep-generative model is to learn parametric class-conditional probability distributions over the intermediate features of discriminative deep networks. Here too, log-likelihoods w.r.t the learnt distributions are evaluated. These methods have a significant advantage that they do not require training a new model, and they can be directly applied to existing pre-trained discriminative models. In such methods, however, an easily tractable form of the density (such as multivariate Normal) is typically assumed. In \cite{lee2018simple}, for instance, the class-conditional distributions are modeled by multivariate Gaussian distributions with shared covariance across all classes (this is known as homoscedasticity). Under such an assumption, the log-likelihood simplifies to the Mahalanobis distance. Further, the densities are learnt on the original high-dimensional Euclidean feature spaces, without any modeling the underlying structure of the features therein. According to the \emph{manifold hypothesis}, however, real-world high-dimensional data lie on low-dimensional manifolds embedded within the high-dimensional space.

We present in this paper, therefore, a principled and modular approach for detecting OOD data in deep neural networks based on subspace techniques and probabilistic modeling of the deep-features within a DNN. 
Specifically, our contributions are as follows:





\begin{enumerate}
	\item We propose the application of linear statistical dimensionality reduction techniques and nonlinear manifold-learning techniques on the high-dimensional features in order to capture the true subspace spanned by the features. We hypothesize that such lower-dimensional feature embeddings can mitigate the \emph{curse of dimensionality}, and improve feature-based methods. 
	
	\item We show that the feature reconstruction error, which is the $L_2$-norm of the difference between the original feature in the high-dimensional space and the pre-image of its low-dimensional reduced embedding, is robust and very effective for OOD detection.
	
	\item We show empirically that the homoscedasticity assumption for class-conditional distributions is not universally valid, and when violated can result in OOD detection performance inferior to even that obtained by Softmax scores. Modeling with more general types of densities in the appropriate subspace can result in improvement in the detection of OOD samples. Our approach, therefore, opens the way for these richer, more complex generative models for the features.   
	
	\item We demonstrate via extensive testing that our approach is modular and proposes a variety of competitive design points and corresponding confidence scores (e.g. linear vs. nonlinear, feature reconstruction error vs. log-likelihood). It outperforms the baseline (up to 13 percentage points in AUROC and AUPR). 
	
\end{enumerate}

\section{Approach}
\label{sec:approach}

The \emph{manifold hypothesis} states that real-world high-dimensional data lie on low-dimensional manifolds embedded within the high-dimensional space. \cite{tenenbaum2000global} prescribes that these high-dimensional spaces should be modeled by appropriate low-dimensional manifolds and sub-spaces. 
In the context of intermediate features of a deep neural network, this implies that the features do not span the entire high-dimensional space they occupy.
This poses significant challenges for feature-based methods. The high dimensionality of the feature space makes it very challenging, often impossible, to perform a variety of otherwise routine tasks on the features, a phenomenon known as the  \emph{curse of dimensionality}. For instance, it leads to rank-deficiency in the data-matrix of the features. 
We postulate that the application of appropriate subspace techniques can successfully mitigate the complications arising from high-dimensional feature spaces. 

\begin{figure}
	\centering
	\includegraphics[width=0.47\textwidth]{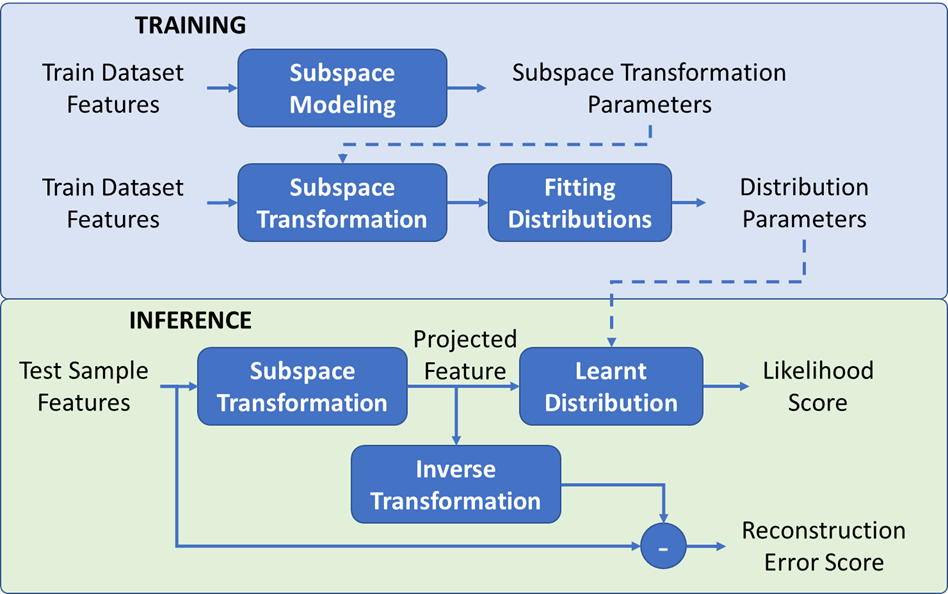}
	\caption{Proposed system.}
	\label{fig:system}
\end{figure}

A flow-diagram of our approach is shown in Figure \ref{fig:system}. The features induced by the training dataset are used to learn the parameters of a transformation that will project the high-dimensional feature onto an appropriate subspace. The parameters of the inverse transformation are also learnt simultaneously. We then fit class-conditional probability densities to the features in the reduced subspace. The parameters of the distributions are estimated by the usual maximum-likelihood approach. During inference, a test sample is projected into the modeled subspace via the learnt subspace transformation, and the likelihood (or log-likelihood) is calculated w.r.t the learnt distribution. The projected sample is inverse-transformed into the original space and a reconstruction-error score is calculated as the $L_2$ norm of the difference between the original and reconstructed vectors. In what follows, we provide details about our solution, and demonstrate its effectiveness.


\subsection{Methodology}

Suppose we have a deep network trained to recognize samples from $N$ classes, $\{C_k\}, k=1, \ldots, N$. Let $f_i(\mb{x})$ denote the output at the $i^{th}$ layer of the network, and $n_i$ its dimension. We then use linear and non-linear techniques to model, respectively, the subspaces or the sub-manifolds on which these features lie. In one class of methods under our approach, an inverse mapping from the reduced subspace to the original feature space is learned and used to compute the pre-image of a reduced feature and calculate the feature reconstruction error. In the second class of methods, we proceed to fit class-conditional probability distributions to the features within these subspaces or sub-manifolds. By fitting distributions to the deep features induced by training samples, we are effectively defining a generative model over the deep feature space. During inference, for both classes of methods, the log-likelihood scores w.r.t. the learnt distributions, and the feature reconstruction errors of a test sample are calculated on a per-layer basis and used to derive confidence scores that can discriminate in-distribution samples from OOD samples. 

\subsection{Subspace Modeling}

\subsubsection{Linear Statistical Dimensionality Reduction:} 
\begin{figure}
	\centering
	\includegraphics[width=0.43\textwidth]{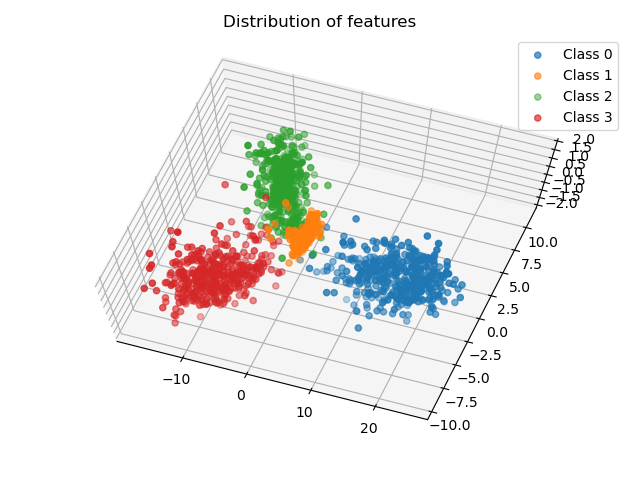}
	\caption{Distribution of features in 3D space. The features for Class 1 have spread mainly in the Z-dimension.}
	\label{fig:dist_3d}
\end{figure}

In the context of fitting probability distributions in high-dimensional feature spaces, the \emph{curse of dimensionality} manifests itself as rank-deficiencies of data matrices, making it impossible to estimate distribution parameters such as the covariance matrices. One way to address this problem consists in using linear statistical dimensionality reduction methods such as principal component analysis (PCA) \cite{10.2307/2333955, jackson2003user}. Table \ref{table:data_ranks} shows the severe rank deficiencies in the higher-dimensional inner layers of a Resnet18 deep network used in our experiments, and how the sparsely-populated high dimensional feature space was meaningfully reduced with PCA. Dimensionality reduction can be applied either on the entire feature set at once, or on a per-class basis. In what follows, we refer to the former as global and the latter as per-class. 
Global dimensionality reduction, which was explored in \cite{ahuja2019probabilistic}, is appealing when the number of samples per class is small relative to the feature dimension, when dealing with large number of classes and in cases where class labels are not available (semi-supervised OOD). However, it may not adequately model the feature space's underlying structure as it does not take advantage of the fact that there can be multiple well-separated clusters, corresponding to separate classes or groups of classes. Applying dimensionality reduction separately to the cluster of each class can often result in much better modeling of that cluster's subspace. The situation is shown in Figure \ref{fig:dist_3d}.

\subsubsection{Nonlinear manifold learning} 
\label{sssec:nonlinear}

PCA is most effective at subspace modeling if the underlying data has a normal distribution, since PCA can only remove second-order dependencies \cite{shlens2014tutorial}. However, the assumption that the features have a normal distribution was justified only at the penultimate layer of a neural network \cite{lee2018simple}. For other layers, the distribution of features could significantly depart from the normal distribution. In general, it will depend on the dataset of images used to induce the feature set, as well as the network topology. In such situations, modeling the data as living in a lower-dimensional submanifold can yield vastly improved outcomes. Here, we use kernel PCA (kPCA) \cite{789} to model the underlying non-linear structure of the data. The choice of kernel PCA is motivated by the fact that it is a nonlinear extension of PCA, which allows for a direct comparison of performance between the linear and nonlinear OOD schemes. In addition, it is computationally less demanding than other manifold learning methods such as Isomap \cite{tenenbaum2000global} and locally linear embedding (LLE) \cite{roweis2000nonlinear}, and also because these other methods can be described as kPCA on specially constructed Gram matrices \cite{ham2004kernel}. 

In essence, existing density modeling techniques make certain assumptions, and on those occasional situations when those assumptions are met, they give quite good results. But more often than not, those assumptions do not hold and in such cases, appropriate modeling of the subspaces can yield substantial improvements in performance.


\subsection{Probabilistic Modeling of deep features:}

From the theory of linear discriminant analysis (LDA), it is known that in a generative classifier in which the underlying class-conditional distributions $p(f_i(\mb{x})|C_k)$ are Gaussians with the same covariance across all classes, the posterior distribution $p(C_k|f_i(\mb{x}))$ is equivalent to the softmax function with linear separation boundaries \cite{bishop2006pattern}. On the basis of this, \cite{lee2018simple} argued that the class-conditional densities of the penultimate layer of a deep-network could be well represented by multivariate Gaussians with same covariance. However, they proceeded to use this model across all layers of the network, not just the penultimate layer. We show empirically that the homoscedastic assumption does not hold true, even in the final layers of a very simple network. We constructed and trained a CNN architecture to classify the first four digits of the MNIST dataset, with the dimension of the penultimate layer set to $3$ for visualization purposes. Figure \ref{fig:dist_3d} clearly shows that the class-conditional feature distributions cannot be modeled with the same covariance. In this work, therefore, we relax the assumption of same covariance, and instead employ the more general parametric probability distribution: separate multivariate Gaussian distribution for each class without the assumption of same covariance. This corresponds to the more general QDA (quadratic discriminant analysis) classifier. Further in this paper, we show elements indicating that this more general Gaussian model remains appropriate only at the penultimate layer. As we move deeper into the network from the output layer, the distributions of features may significantly depart from the normal distribution.

{\bf Estimating parameters:} The parameters of the class-conditional densities are estimated from the training set samples by maximum-likelihood. If the chosen density is a multivariate Gaussian (separate covariance), the maximum-likelihood values of the mean/covariance for class $k$ is given by the unbiased sample mean/covariance:

\be
\label{eq:mean_covar}
\begin{split}
	\mb{\mu}_k &= \frac{1}{M_k}\sum_{\mb{x} \in C_k} f(\mb{\mb{x}}), \\
	\mb{\Sigma}_k &= \frac{1}{M_k}\sum_{\mb{x} \in C_k}\left(f(\mb{x})-\mb{\mu}_k)(f(\mb{x})-\mb{\mu}_k\right)^T
\end{split}
\ee
where $f(\mb{x})$ are the feature values from the network and the layer subscript $i$ has been dropped for notational convenience. If the covariance is assumed to be the same across all classes, then all samples $\mb{x}$ in the training set are used to estimate the covariance, rather than only $\mb{x} \in C_K$. The estimation of the mean remains unchanged. If the chosen density is a GMM, its parameters are estimated using an expectation-maximization (EM) procedure. To choose the number of components in the GMM (i.e. model selection), we adopt the Bayesian Information Criteria (BIC) to penalize complex models; details on EM and BIC can be found in \cite{bishop2006pattern}. 

One of the challenges in using more general distribution types is that the numbers of parameters increases but the number of training samples from which such parameters will be estimated remains the same, e.g. multiple covariance matrices may need to be be estimated from the same amount of training data. In such situations, the assumption of same covariance across classes may seem attractive, since all training samples are available for estimation of a single covariance matrix. However, as we demonstrate, by applying appropriate subspace techniques, we can not only mitigate these issues, but actually improve the eventual OOD detection scores by enabling the use of more general distributions. 

\subsection{Feature Reconstruction Error:}
The log-likelihood score is an effective confidence metric to detect OOD samples, as it gives the distance of the sample w.r.t the learnt distributions in the reduced feature subspace. However, after application of the subspace technique, it is possible for an OOD high-dimensional feature to get projected onto a reduced feature that lies very close to the learnt distributions. This would result in high log-likelihood scores leading to the sample being erroneously interpreted as an in-distribution sample. To handle such situations, we calculate the \emph{feature reconstruction error} and use it as a confidence score for OOD detection. The reconstruction error is the norm of the difference between the original feature vector and the pre-image of its corresponding reduced feature. As the mapping from the high-dimensional feature space to the lower-dimensional reduced subspace is non-injective, there is not a uniquely defined inverse image (pre-image) for a reduced feature. In the linear case, a common practice is to use the Moore-Penrose pseudo-inverse of the forward transformation. In the kPCA case, there are many methods for computing a pre-image, and we refer the readers to \cite{806} for a seminal paper on the subject.

\begin{table}
	\caption{Resnet18 feature dimensions and data-matrix ranks for CIFAR10}
	\label{table:data_ranks}
	\centering
	\begin{tabular}{cccc}
		\toprule
		\textbf{Layer} & Layer 0 & Layer 1 & Layer 2\\
		\cmidrule(r){1-1}\cmidrule(r){2-4}
		Dimension & 512 & 256 & 512 \\
		Rank & 85 & 255 & 478 \\
		With $99.5\%$ PCA & 29 & 239 & 463 \\
		\bottomrule
	\end{tabular}
\end{table}

\begin{table*}
	\caption{OOD detection performance}
	\label{table:auroc}
	\centering
	\begin{tabular}{cccccccccccc}
		
		\toprule
		\midrule
		\multicolumn{2}{c}{CIFAR100} & \multicolumn{5}{c}{SVHN} & \multicolumn{5}{c}{LSUN}  \\
		\cmidrule(r){3-7}\cmidrule(r){8-12}
		\multicolumn{2}{c}{(Wide-Resnet)} & Mahal & LL & PES & K-LL & K-PES &  Mahal & LL & PES & K-LL & K-PES \\
		\cmidrule(r){1-2}\cmidrule(r){3-3}\cmidrule(r){4-5}\cmidrule(r){6-7}\cmidrule(r){8-9}\cmidrule(r){9-10}\cmidrule(r){11-12}
		\multirow{2}{*}{Layer3} & AUROC & 91.5 & 92.8 & 93.1 & 92.9 & \textbf{93.4} & \textbf{98.5} & \textbf{98.8} & 98.3 & \textbf{98.9} & 98.2\\
		& AUPR & 80.5  & 83.1 & 82.9 & 83.5 & \textbf{88.3} & \textbf{98.6} & \textbf{98.9} & 98.2 & \textbf{99.0} & 98.2\\
		\cmidrule(r){1-1}\cmidrule(r){2-2}\cmidrule(r){3-7}\cmidrule(r){8-12}
		\multirow{2}{*}{Layer2} & AUROC & 91.2 & 90.0 & 93.2 & 95.0 & \textbf{95.8} & \textbf{98.7} & \textbf{99.1} & 98.4 & \textbf{98.9} & 98.3\\
		& AUPR & 78.4 & 80.0 & 82.8 & 88.4 & \textbf{90.4} & \textbf{98.8} & \textbf{99.2} & 98.4 & \textbf{99.0} & 98.4\\
		\cmidrule(r){1-1}\cmidrule(r){2-2}\cmidrule(r){3-7}\cmidrule(r){8-12}
		\multirow{2}{*}{Layer1} & AUROC & 75.0 & 84.8 & 79.3 & \textbf{86.0} & 75.9 & \textbf{97.3} & 95.1 & \textbf{97.1} & 94.1 & 91.5\\
		& AUPR & 54.6 & 72.1 & 58.4 & \textbf{76.1} & 55.1 & \textbf{97.5} & 95.6 & \textbf{97.4} & 94.7 & 92.9\\
		\cmidrule(r){1-1}\cmidrule(r){2-2}\cmidrule(r){3-7}\cmidrule(r){8-12}
		\multirow{2}{*}{Layer0} & AUROC & 73.6 & 81.1 & 76.8 & \textbf{83.0} & 75.8 & 90.6 & 89.4 & \textbf{92.8} & \textbf{92.9} & 91.8\\
		& AUPR & 53.0 & 57.7 & 56.6 & \textbf{60.0} & 54.6 & 92.3 & 91.5 & \textbf{93.9} & \textbf{94.0} & 93.1\\
		\cmidrule(r){1-1}\cmidrule(r){2-2}\cmidrule(r){3-7}\cmidrule(r){8-12}
		\multirow{2}{*}{Softmax} & AUROC & \multicolumn{5}{c}{74.3}  & \multicolumn{5}{c}{84.7}\\
		& AUPR & \multicolumn{5}{c}{61.2} & \multicolumn{5}{c}{87.7}\\
		\midrule
		\multicolumn{2}{c}{CIFAR10} & \multicolumn{5}{c}{SVHN} & \multicolumn{5}{c}{LSUN}  \\
		\cmidrule(r){3-7}\cmidrule(r){8-12}
		\multicolumn{2}{c}{(Resnet)} & Mahal & LL & PES & K-LL & K-PES &  Mahal & LL & PES & K-LL & K-PES \\
		\cmidrule(r){1-2}\cmidrule(r){3-3}\cmidrule(r){4-5}\cmidrule(r){6-7}\cmidrule(r){8-9}\cmidrule(r){9-10}\cmidrule(r){11-12}
		\multirow{2}{*}{Layer2} & AUROC & 94.6 & 94.5 & 77.2 & \textbf{98.9} & 98.5 & 98.8 & \textbf{99.4} & 95.3 & 98.6 & \textbf{99.0}\\
		& AUPR & 88.8 & 90.3 & 55.5 & \textbf{97.6} & 96.1 & 98.8 & \textbf{99.4} & 94.7 & 98.8 & \textbf{99.1}\\
		\cmidrule(r){1-1}\cmidrule(r){2-2}\cmidrule(r){3-7}\cmidrule(r){8-12}
		\multirow{2}{*}{Layer1} & AUROC & 86.4 & 88.8 & 48.5 & \textbf{92.7} & 92.4 & 72.5 & 86.0 & 65.2 & 83.3 & \textbf{87.0}\\
		& AUPR & 71.5 & 78.2 & 26.7 & \textbf{84.6} & 82.2 & 73.1 & 86.1 & 65.6 & 84.4 & \textbf{87.7}\\
		\cmidrule(r){1-1}\cmidrule(r){2-2}\cmidrule(r){3-7}\cmidrule(r){8-12}
		\multirow{2}{*}{Layer0} & AUROC & 95.2 & 95.0 & \textbf{96.7} & 94.6 & 93.9 & \textbf{95.1} & \textbf{95.5} & \textbf{95.3} & \textbf{95.6} & \textbf{95.1}\\
		& AUPR & 92.9 & 93.1 & \textbf{94.5} & 92.6 & 91.1 & \textbf{95.9} & \textbf{96.1} & \textbf{96.1} & \textbf{95.9} & \textbf{96.0}\\
		\cmidrule(r){1-1}\cmidrule(r){2-2}\cmidrule(r){3-7}\cmidrule(r){8-12}
		\multirow{2}{*}{Softmax} & AUROC & \multicolumn{5}{c}{93.4}  & \multicolumn{5}{c}{94.0}\\
		& AUPR & \multicolumn{5}{c}{88.7} & \multicolumn{5}{c}{94.8}\\
		\midrule 
		\multicolumn{2}{c}{SVHN} & \multicolumn{5}{c}{CIFAR10} & \multicolumn{5}{c}{LSUN}  \\
		\cmidrule(r){3-7}\cmidrule(r){8-12}
		\multicolumn{2}{c}{(Resnet20)} & Mahal & LL & PES & K-LL & K-PES &  Mahal & LL & PES & K-LL & K-PES \\
		\cmidrule(r){1-2}\cmidrule(r){3-3}\cmidrule(r){4-5}\cmidrule(r){6-7}\cmidrule(r){8-9}\cmidrule(r){9-10}\cmidrule(r){11-12}
		\multirow{2}{*}{Layer2} & AUROC & \textbf{94.2} & \textbf{93.8} & 85.2 & 93.6 & 93.7 & \textbf{94.3} & \textbf{93.9} & 90.1 & \textbf{93.8} & 93.5\\
		& AUPR & \textbf{91.9} & \textbf{91.5} & 66.7 & 91.4 & 91.1 & \textbf{97.8} & \textbf{97.7} & 93.9 & \textbf{97.7} & 97.5 \\
		\cmidrule(r){1-1}\cmidrule(r){2-2}\cmidrule(r){3-7}\cmidrule(r){8-12}
		\multirow{2}{*}{Layer1} & AUROC & 90.4 & 94.9 & 94.2 & \textbf{96.3} & 94.7 & 90.6 & 95.2 & 94.5 & \textbf{96.9} & 94.9\\
		& AUPR & 87.8 & 92.3 & 92.0 & \textbf{94.5} & 92.6 & 96.5 & 98.1 & 97.9 & \textbf{98.8} & 98.0 \\
		
		\cmidrule(r){1-1}\cmidrule(r){2-2}\cmidrule(r){3-7}\cmidrule(r){8-12}
		\multirow{2}{*}{Layer0} & AUROC & 92.3 & \textbf{96.8} & 96.0 & \textbf{96.7} & 95.6 & 92.5 & \textbf{97.1} & 96.0 & 97.0 & 95.8 \\
		& AUPR & 89.6 & \textbf{95.7} & 94.6 & \textbf{95.6} & 94.1 & 97.2 & \textbf{98.9} & 98.5 & 98.9 & 98.4 \\
		\cmidrule(r){1-1}\cmidrule(r){2-2}\cmidrule(r){3-7}\cmidrule(r){8-12}
		\multirow{2}{*}{Softmax} & AUROC & \multicolumn{5}{c}{93.0}  & \multicolumn{5}{c}{92.5}\\
		& AUPR & \multicolumn{5}{c}{89.9} & \multicolumn{5}{c}{96.9}\\
		\midrule
		\bottomrule
	\end{tabular}
\end{table*}

\section{Experiments and Results}
\label{sec:Results}

\paragraph{Experimental setup and evaluation metrics} For comparison, following other papers, we use CIFAR10, CIFAR100 \cite{krizhevsky2009learning}, and SVHN \cite{netzer2011reading} as the in-distribution datasets. To ensure that we test across networks of various complexities, we train SVHN on Resnet20 ($0.27$M trainable parameters), CIFAR10 on Resnet18 ($11.2$M parameters) \cite{He2016DeepRL}, and CIFAR100 on Wide-Resnet ($36.5$M parameters) \cite{zagoruyko2016wide}. For CIFAR10 and CIFAR100, we use SVHN dataset and a resized version of the LSUN datasets \cite{yu15lsun} as the OOD datasets. For SVHN as in-distribution dataset, we use CIFAR10 and LSUN as OOD datasets. In all experiments, the parameters of the fitted density functions are estimated from the training split of the in-distribution dataset, while performance metrics (accuracy, AUPR, AUROC) are calculated on the test split.  As described in Section \ref{sssec:nonlinear}, the distribution of features depends on the in-distribution dataset and network topology, and we apply both the linear (PCA) and nonlinear (kPCA with RBF kernel) subspace techniques to model the feature subspace. Determining the best subspace representation requires selecting hyper-parameters for PCA (retained data variability), and kPCA (reduction level, and gamma factor). In our experiments, these were determined with the help of a hold-out auxiliary OOD dataset entirely different from the test data. Specifically, we used Tiny ImageNet \cite{le2015tiny} as the hold-out auxiliary dataset. Note that this is different from Outlier Exposure (OE) \cite{hendrycks2019oe}), where such a dataset was used to actually train the network. We don't modify the network training in any way; the hold-out auxiliary dataset is used solely for hyper-parameter selection during subspace modeling. We also note that we performed a rather coarse search over the hyper-parameter space; nonetheless, we found that the selected hyper-parameters generalized well and gave good performance on the unseen test sets. It is quite likely that a more exhaustive hyper-parameter search would have yielded even better outcomes than we currently report. Additionally, we perform our experiments on three layers of the networks listed above, with layers chosen to be located uniformly along the network path. Our framework can be easily extended to add any layer. The layers are labelled as 0, 1, and 2, with 0 being the outermost layer, and 1, 2 being progressively deeper within the network. Since WideResNet is a bigger and more complex network, we tap an additional layer (Layer 3).
More details about the network topologies, which layers are modeled, and the hyper-parameters are provided in the supplementary material.

\begin{figure}
	\centering
	\includegraphics[width=0.43\textwidth]{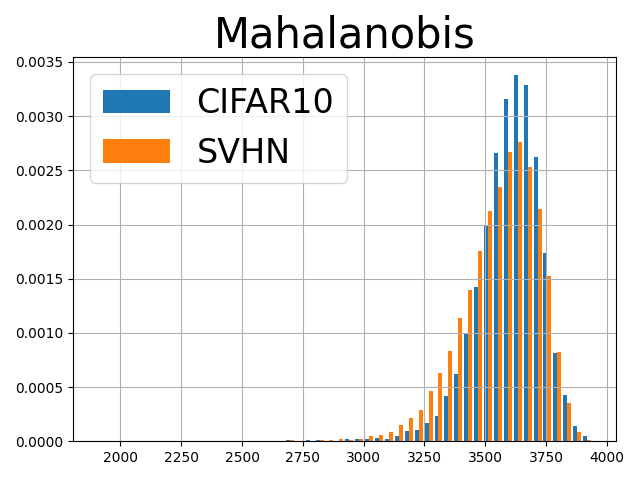}
	\includegraphics[width=0.43\textwidth]{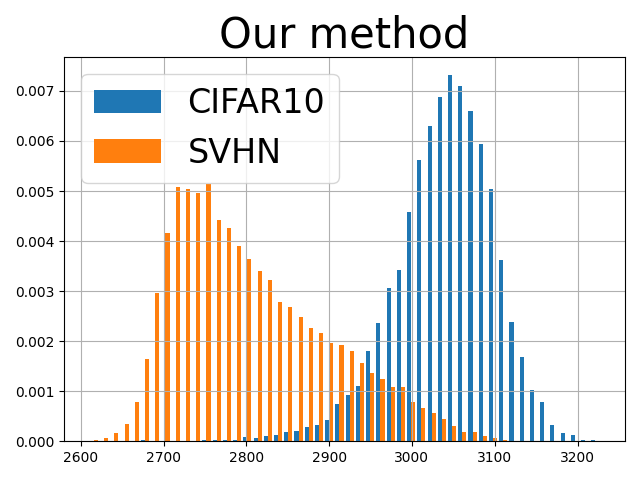}
	\caption{Histogram of scores for in-distribution (CIFAR10) and out-of-distribution data (SVHN). Left panel is Mahalanobis; right panel is our method (log-likelihood in KPCA-reduced subspace)}
	\label{fig:histograms}
\end{figure}

During testing, we use both the log-likelihood scores as well as the feature reconstruction error scores to distinguish between in distribution and out-of-distribution data, effectively creating binary classifiers. The performance of these is characterized by typical methods such as the precision-recall (PR) curve or the receiver operating characteristics (ROC) curve, and we report the area under both types of curves (AUROC and AUPR).

Finally, note that although we have investigated the improvement in OOD detection performance on a per-layer basis, it is possible to combine all such per-layer scores in order to improve the overall detection performance. For instance, the scores could be combined using any binary classifier such as logistic regression trained using a hold-out auxiliary OOD data set. It is also worth noting that the log-likelihood scores and reconstruction-error scores are complementary by design. The former measures the OOD detection performance in the learnt low-dimensional subspace, while the latter measures the distance of the feature vector from the learnt subspace. We are currently working to combine these two scores to yield a unique confidence score.

\paragraph{Subspace modeling} In Table \ref{table:data_ranks}, we show the ranks of the data-matrices at various layers in Resnet18 trained on CIFAR10. It is seen that the rank is much lower than the feature-dimension at that layer, indicating that the features actually reside in a lower-dimensional subspace and hence the need for appropriate subspace modeling before fitting distributions. An intuition for the extent of dimensionality reduction possible can be obtained by applying PCA when retaining a very high amount of data variability. It is seen, for instance, that for Layer 0, the subspace dimension can be at most $85$ even though the original feature dimension is $512$. Further, applying PCA with $99.5$\% variability retention reduces its dimension to $29$, indicating the $99.5$\% of the information in the $512$-dimensional features is actually contained within a $29$-dimensional subspace! The results for other layers are less dramatic but nonetheless point to the need for appropriate subspace modeling. The corresponding results for other datasets and networks show similar behavior, and are provided in the supplementary material.

\begin{figure}
	\centering
	\includegraphics[width=0.23\textwidth]{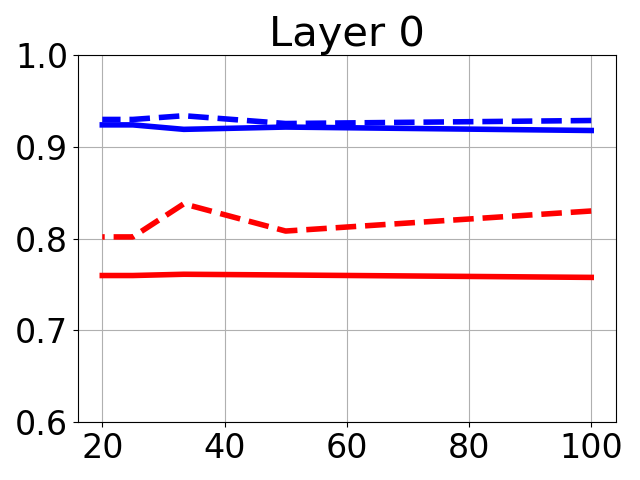}
	\includegraphics[width=0.23\textwidth]{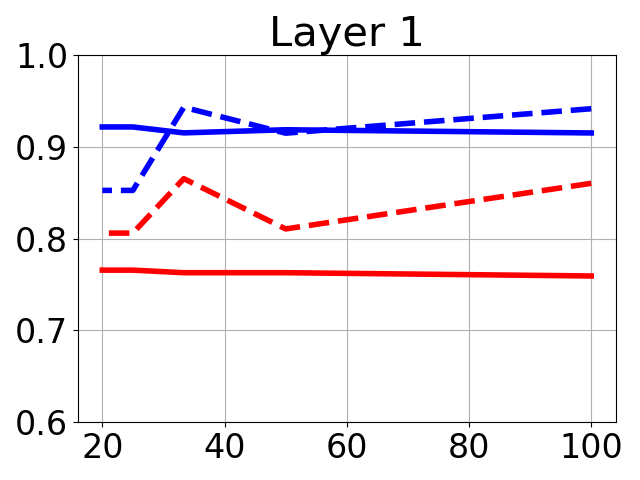}
	\includegraphics[width=0.23\textwidth]{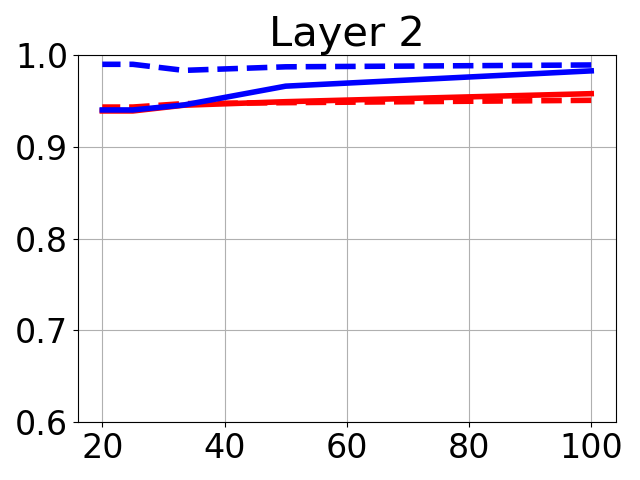}
	\includegraphics[width=0.23\textwidth]{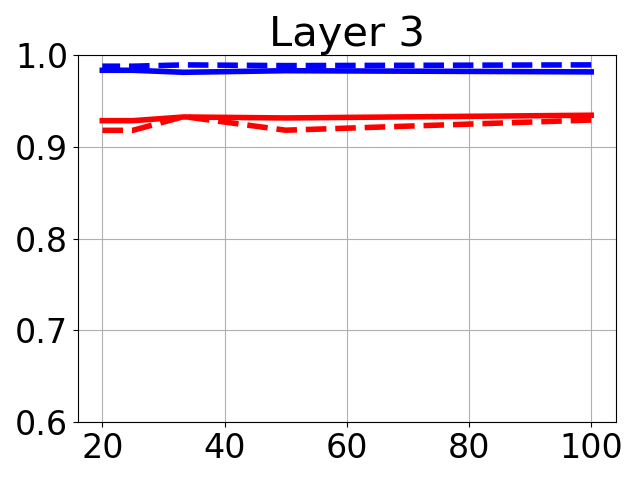}
	\caption{AUROC (Y-axis) for OOD detection with CIFAR100 (in-distribution) and SVHN (red) and LSUN (blue) as OOD sets as we decrease the percentage of training data used (X-axis) for our OOD method all the down to 20\%. Solid line shows results for the reconstruction-error scores, and the dotted line for the log-likelihood scores.}
	\label{fig:robustness}
\end{figure}

\paragraph{OOD detection results} Next, we report the OOD detection performance using AUPR and AUROC scores in Tables \ref{table:auroc}. For baseline comparison, we use the output Softmax scores, as well as Mahalanobis scores which corresponds to modeling in the original feature space with multivariate Gaussians with the same covariance across classes. The Mahalanobis baseline  requires average pooling to be applied on the features being tapped and its performance varies significantly when different pooling levels are applied. We report here the best results we can obtain for this baseline. With different pooling settings, the improvements provided by our approach are much larger (see Fig. \ref{fig:histograms}). For a given dataset and network, we deem any set of scores for which both the AUPR and AUROC results are within 
a half-percentage point of each other to be a statistical tie. As seen from Table \ref{table:auroc}, the OOD detection results obtained by our method - both with linear and non-linear subspace modeling - either exceeds the baselines by large margins or are a virtual tie. We also observe that, in addition to the log-likehood scores, the feature reconstruction error scores are highly effective at discriminating OOD data. In a couple of instances, the feature reconstruction error from linear subspace modeling gives rather poor performance. 
In such instances, nonlinear subspace modeling gives substantially improved results. In general, we notice a trend of the nonlinear scheme providing better results than its linear counterpart as we progress deeper into the network. As previously discussed, the nature of the feature space is dependent on the network topology and the training images used to induce it. In our observations, the outer-most layers produce features with a certain degree of regularity while deeper layers have feature spaces that may exhibit complex nonlinearity in their structure. Also, note that while Mahalanobis scores often work well, there are quite a few instances (e.g. Layer 1 results for CIFAR10) in which it under-performs even the Softmax scores. This happens particularly in some of the deeper layers in which the homoscedasticity assumption is more likely to be violated. 

\paragraph{Robustness to reduced data} In practical situations, we might not have access to the entire training dataset. We show that our method remains very effective at OOD detection even when its training (subspace modeling and distribution fitting) is performed with a fraction of the training data. The plots in Figure \ref{fig:robustness} show the variations in AUROC for the CIFAR100 dataset, for the log-likelihood and feature reconstruction error scores, using both linear and non-linear subspace modeling, as the percentage for training data is gradually reduced to $20$\%. We see that the performance remains very stable, showing a decrease of less than $1$\% in AUROC scores in the majority of cases. On a very few occasions, it drops by over $5$\%, which is actually impressive given that the amount of training data used is only a fifth of the total size available. We observe this trend for the other datasets as well and the corresponding plots for those are provided in the supplementary material.


\section{Conclusions and Future Work}
\label{sec:conclusions}
This paper presented a principled and modular approach for detecting out-of-distributions samples with the use of subspace techniques and probabilistic modeling of features. We showed that the sparsely-populated high-dimensional feature space can be meaningfully reduced using appropriate linear and nonlinear subspace techniques. Beyond just out-of-distribution, we hypothesized that any feature-based method would benefit from it. We also showed that the feature reconstruction error can be used very effectively as a confidence score for OOD detection.
The methodology was theoretically motivated, and experimentally proven by showing improvements to out-of-distribution detection on image data using popular network architectures. 
Future work will seek to combine the reconstruction error with the log-likelihood score for improved performance.

\bibliography{DFM_AAAI_2021}

\end{document}